\documentclass[conference]{IEEEtran}
\IEEEoverridecommandlockouts
\usepackage{cite}
\usepackage{amsmath,amssymb,amsfonts}
\usepackage{algorithmic}
\usepackage{graphicx}
\usepackage{textcomp}
\usepackage{xcolor}

\usepackage{booktabs}
\usepackage{amssymb}
\def\BibTeX{{\rm B\kern-.05em{\sc i\kern-.025em b}\kern-.08em
    T\kern-.1667em\lower.7ex\hbox{E}\kern-.125emX}}

\begin{document}

\title{Object Placement for Anything}

\author{\IEEEauthorblockN{Bingjie Gao}
\IEEEauthorblockA{\textit{Shanghai Jiao Tong University} \\
Shanghai, China \\
whynothaha@sjtu.edu.cn}
\and
\IEEEauthorblockN{Bo Zhang}
\IEEEauthorblockA{\textit{Shanghai Jiao Tong University} \\
Shanghai, China \\
bo-zhang@sjtu.edu.cn}
\and
\IEEEauthorblockN{Li Niu$^{*}$}
\IEEEauthorblockA{\textit{Shanghai Jiao Tong University} \\
Shanghai, China \\
ustcnewly@sjtu.edu.cn}
}

\maketitle
\renewcommand{\footnoterule}{\noindent\rule{0.35\columnwidth}{0.4pt}\vspace{2pt}}
\renewcommand{\thefootnote}{}
\footnotetext{*Corresponding author.}

\begin{abstract}
Object placement aims to determine the appropriate placement (\emph{e.g.}, location and size) of a foreground object when placing it on the background image.  Most previous works are limited by small-scale labeled dataset, which hinders the real-world application of object placement. In this work, we devise a semi-supervised framework which can exploit large-scale unlabeled dataset to promote the generalization ability of discriminative object placement models.
   The discriminative models predict the rationality label for each foreground placement given a foreground-background pair. To better leverage the labeled data, under the semi-supervised framework, we further propose to transfer the knowledge of rationality variation, \emph{i.e.}, whether the change of foreground placement would result in the change of rationality label, from labeled data to unlabeled data. 
   Extensive experiments demonstrate that our framework can effectively enhance the generalization ability of discriminative object placement models.
\end{abstract}

\begin{IEEEkeywords}
object placement, semi-supervised learning
\end{IEEEkeywords}

\section{Introduction}
\label{sec:intro}
Image composition \cite{Niu_Cong_Liu_Yan_Zhang_Liang_Zhang_2021} aims to produce a realistic composite image for a foreground-background pair, which can be widely applied in data augmentation \cite{Remez_Huang_Brown_2018}, artistic creation \cite{Tripathi_Chandra_Agrawal_Tyagi_Rehg_Chari_2019}, and so on. As a subtask of image composition, object placement \cite{Zhang_Meng_Liu_Nie_Zhong_Fan_Ji,Zhu_Lin_Cohen_Kuen_Zhang_Chen_2023,Zhu_Lin_Cohen_Kuen_Zhang_Chen,Zhou_Liu_Niu_Zhang_2022} focuses on finding the appropriate placement 
 (\emph{e.g.}, size, location) for foreground objects. Learning object placement is a challenging task because various factors (\emph{e.g.}, actual size, depth, semantics) should be taken into account.

Object placement methods can be divided into generative methods \cite{Zhang_Meng_Liu_Nie_Zhong_Fan_Ji,Zhou_Liu_Niu_Zhang_2022,Zhang_Wen_Min_Wang_Han_Shi_2020,Zhu_Lin_Cohen_Kuen_Zhang_Chen} and discriminative methods \cite{Liu_Zhang_Li_Niu_Liu_Zhang_2021,Niu_Liu_Liu_Li_2022,Zhu_Lin_Cohen_Kuen_Zhang_Chen_2023}. Generative methods aim to predict one or multiple plausible placements for a foreground-background pair. Specifically, some methods \cite{Zhang_Wen_Min_Wang_Han_Shi_2020,Zhou_Liu_Niu_Zhang_2022} combine foreground and background features with random vectors to predict multiple bounding boxes. Zhang \emph{et al.} \cite{Zhang_Meng_Liu_Nie_Zhong_Fan_Ji} integrate reinforcement learning for dynamic object placement. According to the analysis in \cite{Niu_Liu_Liu_Li_2022}, generative methods performs poorly and unsteadily compared with discriminative methods. 
Discriminative methods focus on assessing the rationality score of each placement for a foreground-background pair. Liu \emph{et al.} \cite{Liu_Zhang_Li_Niu_Liu_Zhang_2021} deem object placement as binary classification task and evaluate the rationality of composite images. However, this method is very slow because all composite images need to pass through the binary classifier.  
Some more efficient discriminative methods \cite{Zhu_Lin_Cohen_Kuen_Zhang_Chen_2023,Niu_Liu_Liu_Li_2022} output a heatmap containing the rationality scores of all locations and scales for a foreground-background pair. The rationality heatmap is produced based on the rationality features of all placements, which are obtained through the interaction between foreground and background. 

Most previous object placement methods \cite{Niu_Liu_Liu_Li_2022,Zhou_Liu_Niu_Zhang_2022,Zhang_Meng_Liu_Nie_Zhong_Fan_Ji} are trained on Object Placement Assessment (OPA) \cite{Liu_Zhang_Li_Niu_Liu_Zhang_2021} dataset, which is a labeled dataset with manually annotated rationality labels for partial placements. 
However, OPA dataset only has a limited number of foreground categories and background scenes, which hinders the real-world application of the models trained on OPA. One obstacle of extending object placement to a wider range of foreground categories and background scenes is the high annotation cost \cite{Liu_Zhang_Li_Niu_Liu_Zhang_2021}. To address this issue, TopNet \cite{Zhu_Lin_Cohen_Kuen_Zhang_Chen_2023} utilizes a large-scale unlabeled dataset, viewing the original placements of foreground objects as positive ones. However, TopNet solely relies on the positive placements, overlooking the annotated negative placements in the existing dataset~\cite{Liu_Zhang_Li_Niu_Liu_Zhang_2021}. 

In this work, we propose a semi-supervised framework for object placement, which can jointly utilize labeled and unlabeled data. 
The labeled data are from small-scale OPA dataset and the unlabeled data are from large-scale Open Images dataset~\cite{Kuznetsova_Rom_Alldrin_Uijlings_Krasin_Pont_Tuset_Kamali_Popov_Malloci_Kolesnikov}. The unlabeled data have a wider range of foreground categories and background scenes, which breaks the constraints of limited labeled data. Our framework performs two steps iteratively: model training and label correction. In the step of model training, we train the object placement model using the labeled data with ground-truth labels and the unlabeled data with pseudo labels. In the step of label correction, we use the latest object placement model to correct the pseudo labels of unlabeled data.
\begin{figure}[tb]
  \centering
\includegraphics[width=0.45\textwidth]{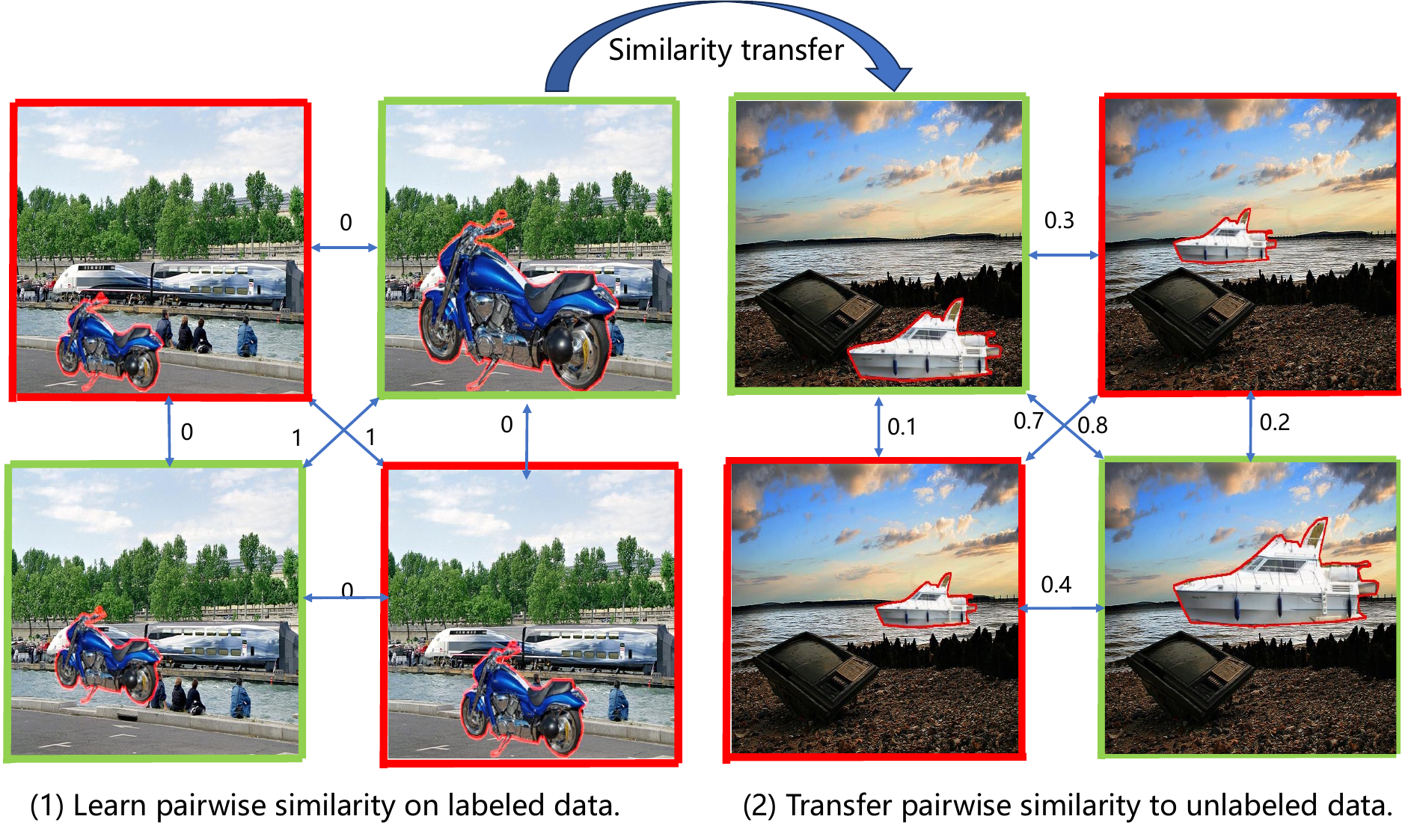}
  \caption{The illustration of similarity transfer from labeled data to unlabeled data. The foregrounds are marked with red outlines. The red and green image borders denote rational and irrational placements, respectively. The numbers above the arrows indicate pairwise similarities of two placements, \emph{i.e.}, whether they have the same rationality label.}
  \label{fig:overview}
\end{figure}

To better leverage the labeled data, we propose to transfer the knowledge of rationality variation from labeled data to unlabeled data, that is, whether the change of foreground placement would result in the change of rationality label given a foreground-background pair. For example, as shown in Fig.~\ref{fig:overview}, when placing a motorcycle on the background, moving the motorcycle from land to water would change the rationality label from 1 to 0, pushing the model to mind the boundary between land and water. When adjusting the motorcycle scale from small to large, the rationality label would change from 1 to 0, or the other way around, which can help the model understand the reasonable range of object scale. Such knowledge is transferrable across different object categories and different background scenes.  
As mentioned before, discriminative object placement models predict the rationality label for each placement based on the rationality feature of this placement. To capture the rationality variation caused by placement variation, we concatenate the rationality features of two placements to predict the similarity label, that is, whether two placements have the same rationality label. This similarity classifier is shared by labeled data and unlabeled data, to facilitate their mutual knowledge transfer. In this paper, we also refer to rationality variation transfer as similarity transfer. 

We conduct extensive experiments on OPA and Open Images datasets, which demonstrates that the discriminative object placement models could benefit from our framework and knowledge transfer. Our main contributions are summarized as follows: 1) We develop a semi-supervised object placement framework which can jointly use labeled data and unlabeled data. 2) Under the semi-supervised framework, we propose to transfer the knowledge of rationality variation from labeled data to unlabeled data. 3) Extensive experiments demonstrate that our framework can significantly promote the generalization ability of discriminative object placement models.


\section{Related Work}
\subsection{Object Placement}
Some traditional object placement methods \cite{Remez_Huang_Brown_2018,Fang_Sun_Wang_Gou_Li_Lu_2019} designed rigid rules governing the determination of feasible locations and sizes for foreground objects. However, the rigid rules may not be applicable to complex scenes. 
Recently, deep learning based methods \cite{Zhu_Lin_Cohen_Kuen_Zhang_Chen_2023,Tripathi_Chandra_Agrawal_Tyagi_Rehg_Chari_2019,Zhang_Wen_Min_Wang_Han_Shi_2020,Liu_Zhang_Li_Niu_Liu_Zhang_2021,Zhang_Meng_Liu_Nie_Zhong_Fan_Ji,liu2024conditional} tend to predict the bounding box or transformation parameters of the foreground object. More recently, discriminative methods have emerged to predict the rationality scores for all placements, achieving more competitive results than previous methods. For instance, Zhu \emph{et al.} \cite{Zhu_Lin_Cohen_Kuen_Zhang_Chen_2023} proposed a transformer-based model named TopNet for real-world compositing. TopNet learns the correlation between foregrounds and backgrounds, producing a 3D heatmap containing the rationality scores of all locations and scales. Most of the above works are limited by small-scale datasets. Zhu \emph{et al.} is the first to leverage large-scale unlabeled data for object placement learning. However, Zhu \emph{et al.} did not take full advantage of existing labeled dataset. In contrast, we construct a semi-supervised framework which can jointly use labeled data and unlabeled data. 

\subsection{Semi-Supervised Learning}
Various methods based on pseudo labeling \cite{Rizve_Duarte_Rawat_Shah_2021} and consistency regularization \cite{Sajjadi_Javanmardi_Tasdizen_2016} have led to the impressive success of Semi-supervised Learning (SSL). The combination of pseudo labeling and consistency regularization has yielded promising outcomes \cite{Zhang_Wang_Hou_Wu_Wang_Okumura_Shinozaki_2021}. Semi-supervised domain adaptation (SSDA) \cite{Li_Li_Shi_Yu_2021} aims to develop a well-performed model that excels in the target domain by leveraging fully labeled source samples and scarcely labeled target samples, along with unlabeled target samples. Current SSDA \cite{Yu_Lin_2023,he2024enhancing} methods typically aim to align target data with labeled source data through feature space mapping and pseudo-label assignment. However, this source-centric approach may incorrectly align target data to the wrong categories of the source data, compromising the classification performance.

\section{Our Method}
\subsection{Semi-supervised Framework} \label{sec:framework}

\begin{figure*}[tb]
  \centering
  \includegraphics[width=0.90\textwidth]{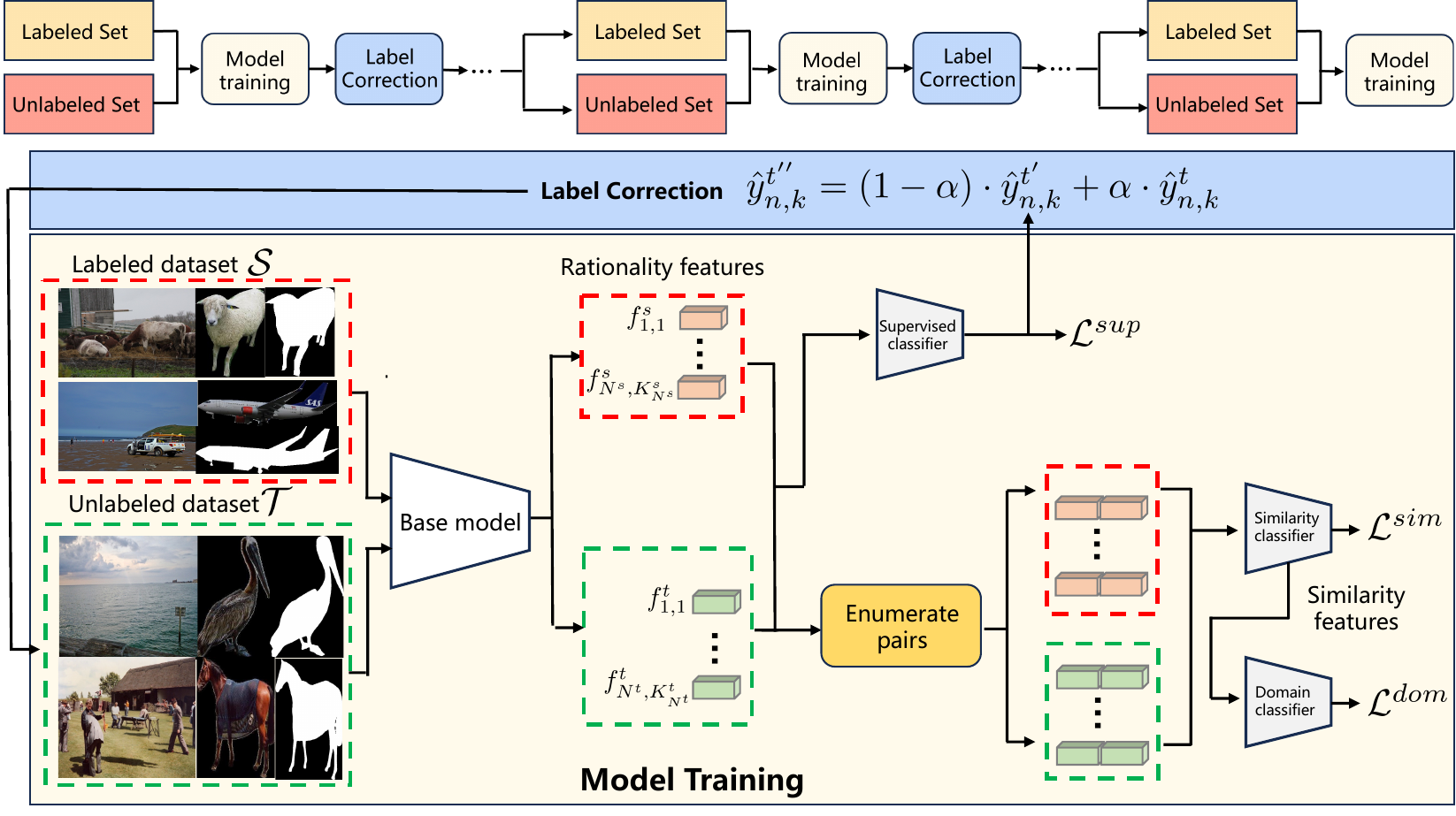}
  \caption{Illustration of our semi-supervised object placement framework, which performs model training and label correction iteratively. In the step of model training, we use base object placement model to extract rationality features, which are sent to the supervised classifier. Pairs of rationality features are sent to similarity classifier and the intermediate similarity features are sent to domain classifier.}
  \label{fig:framework}
\end{figure*}

Given a background image $B$ and a foreground object $F$, discriminative object placement models predict the rationality score $y$ for each foreground placement (\emph{e.g.}, bounding box $b$), indicating whether it is plausible to place the foreground object $F$ within the bounding box $b$ on the background image $B$. 
The bounding box $b$ can be decomposed into the foreground scale $[w, h]$ and the center location $[x, y]$ of foreground on the background image. Thus, one placement (bounding box) corresponds to a foreground scale and a location on the background. 
To predict the rationality score for each placement, the discriminative object placement models produce rationality feature $f$ for each placement, which is obtained by integrating the feature vector of scaled foreground with the pixel-wise background feature vector at certain location. The rationality feature for each placement is responsible for predicting the rationality score of this placement. 

In this work, we jointly use labeled dataset and unlabeled dataset. The labeled dataset is represented by $\mathcal{S} = \{s_n|_{n=1}^{N^s} \}$, in which $N^s$ is the number of foreground-background pairs and $s_n$ contains the required information for the $n$-th foreground-background pair. Specifically, $s_n = (B^s_n, F^s_n, \{ b^s_{n,k} |_{k=1}^{K^s_n} \}, \{y^s_{n,k}|_{k=1}^{K^s_n} \})$, in which $B^s_n$ is the background image, $F^s_n$ is the foreground object, $b^s_{n,k}$ is the bounding box for the $k$-th placement, $y^s_{n,k}$ is the rationality label for the $k$-th placement, $K^s_n$ is the number of annotated placements for the $n$-th foreground-background pair. 

Similarly, the unlabeled dataset is represented by $\mathcal{T} = \{t_n|_{n=1}^{N^t} \}$, in which $t_n = (B^t_n, F^t_n, \{ b^t_{n,k} |_{k=1}^{K^t_n} \}, \{\hat{y}^t_{n,k}|_{k=1}^{K^t_n} \})$. Since unlabeled dataset does not have annotated placements, we randomly sample $K^t_n$ placements $\{ b^t_{n, k} |_{k=1}^{K^t_n}\}$ and denote their pseudo rationality labels as $\{\hat{y}^t_{n, k}|_{k=1}^{K^t_n} \}$.

Our semi-supervised framework performs model training and label correction iteratively. In the step of model training, we train the object placement model using both $\mathcal{S}$ and $\mathcal{T}$. In the step of label correction, we use the latest object placement model to correct the pseudo labels $\hat{y}^t_{n,k}$ in $\mathcal{T}$. In the first iteration, we use off-the-shelf object placement model to acquire the pseudo labels. Specifically, we predict the rationality score $p^t_{n,k}$ for the $k$-th placement using the pretrained object placement model and get the pseudo rationality label $\hat{y}^t_{n, k} = \delta(p^t_{n, k}>\gamma)$, in which $\delta(\cdot)$ is an indicator function and $\gamma$ is a threshold. In the remaining iterations, we use label correction in Section~\ref{sec:label_correction} to update the pseudo rationality labels.

\subsection{Model Training}
\label{model_training}

\subsubsection{Supervised Loss}

We divide discriminative object placement model into two components: base model $\theta^{opa}$ to extract rationality features and supervised classifier $\theta^{sup}$ to predict rationality scores. 
Based on the ground-truth labels $y^s_{n,k}$ in the labeled dataset $\mathcal{S}$ and the pseudo labels $\hat{y}^t_{n,k}$ in the unlabeled dataset $\mathcal{T}$,
we employ the standard binary cross-entropy (BCE) loss $\mathcal{L}^{bce}$:
\begin{equation}
    \mathcal{L}^{sup} = \sum_{n=1}^{N^s} \sum_{k=1}^{K_n} \mathcal{L}^{bce}(y^s_{n,k}, p^s_{n,k}) +  \sum_{n=1}^{N^t} \sum_{k=1}^{K_n} \mathcal{L}^{bce}(\hat{y}^t_{n,k}, p^t_{n,k}),
\end{equation}
in which $p^s_{n,k}$ (\emph{resp.}, $p^t_{n,k}$) is the predicted rationality score for the $k$-th placement given the $n$-th foreground-background pair in $\mathcal{S}$  (\emph{resp.}, $\mathcal{T}$).  

\subsubsection{Similarity Loss}

To transfer the knowledge of rationality variation from labeled dataset $\mathcal{S}$ to unlabeled dataset $\mathcal{T}$, we introduce similarity classifier to judge whether two placements have the same rationality label. As mentioned in Section~\ref{sec:framework}, the discriminative object placement models produce one rationality feature for each placement, which is used to predict the rationality score of this placement. To verify whether two placements have the same rationality label, we concatenate two rationality features corresponding to two placements, which are passed through a similarity classifier $\theta^{sim}$ to predict the similarity score. 

By taking the labeled dataset $\mathcal{S}$ as an example, we denote the rationality features for the $n$-th foreground-background pair as $\{f^s_{n,k}|_{k=1}^{K^s_n}\}$. We concatenate two rationality features $[f^s_{n,i}, f^s_{n,j}]$, which is projected to similarity feature $f^{s,sim}_{n,i,j}$  to predict the similarity score $p^{s,sim}_{n,i,j}$, which is supervised by the ground-truth similarity label $y^{s,sim}_{n,i,j}=\delta(y^{s}_{n,i}=y^{s}_{n,j})$. In analogy, for the unlabeled dataset $\mathcal{T}$, we predict the similarity score $p^{t,sim}_{n,i,j}$, which is supervised by the pseudo similarity label $\hat{y}^{t,sim}_{n,i,j}=\delta(\hat{y}^{t}_{n,i}=\hat{y}^{t}_{n,j})$. We also use BCE loss to supervise the predicted similarity scores:
\begin{eqnarray} \label{Similarity_loss}
    \mathcal{L}^{sim} = \!\!\!\!\!\!\! && \sum_{n=1}^{N^s}\sum_{i\neq j}^{K^s_n}  \mathcal{L}^{bce}(y^{s,sim}_{n,i,j}, p^{s,sim}_{n,i,j}) \nonumber\\
    && + \sum_{n=1}^{N^t}\sum_{i\neq j}^{K^t_n}  \mathcal{L}^{bce}(\hat{y}^{t,sim}_{n,i,j}, p^{t,sim}_{n,i,j}).
\end{eqnarray}

\subsubsection{Domain Loss}
Considering that the labeled dataset $\mathcal{S}$ and the unlabeled dataset $\mathcal{T}$ have considerably different data distributions, which may affect the knowledge transfer from $\mathcal{S}$ to $\mathcal{T}$. Following the terminology in domain adaptation, we refer to $\mathcal{S}$ and $\mathcal{T}$ as two domains.  
To facilitate knowledge transfer, we employ an adversarial loss to reduce the domain gap of similarity features between $\mathcal{S}$ and $\mathcal{T}$. 

We feed the similarity features $f^{s,sim}_{n,i,j}$ in $\mathcal{S}$ and $f^{t,sim}_{n,i,j}$ in $\mathcal{T}$ to the domain classifier $\theta^{dom}$ to predict the domain label ($1$ for $\mathcal{S}$ and $0$ for $\mathcal{T}$). After denoting the ground-truth domain labels as $\{y^{s,dom}_{n,i,j}=1, y^{t,dom}_{n,i,j}=0\}$ and the domain classifier outputs as $\{p^{s,dom}_{n,i,j}, p^{t,dom}_{n,i,j}\}$, the domain classification loss can be written as
\begin{eqnarray} \label{Domain_loss}
    \mathcal{L}^{dom} = \!\!\!\!\!\!\! && \sum_{n=1}^{N^s}\sum_{i\neq j}^{K^s_n}  \mathcal{L}^{bce}(y^{s,dom}_{n,i,j}, p^{s,dom}_{n,i,j}) \nonumber\\
    && + \sum_{n=1}^{N^t}\sum_{i\neq j}^{K^t_n}  \mathcal{L}^{bce}(\hat{y}^{t,dom}_{n,i,j}, p^{t,dom}_{n,i,j}).
\end{eqnarray}
The domain classifier and the other modules are trained in an adversarial learning manner. 
The domain classifier $\theta^{dom}$ is updated to minimize the domain loss $\mathcal{L}^{dom}$, while the rest of modules $\{\theta^{opa}, \theta^{sup}, \theta^{sim} \}$ are optimized to fool the domain classifier to pull close $\mathcal{S}$ and $\mathcal{T}$.

During model training, we jointly optimize the supervised loss, similarity loss, and domain loss:
\begin{equation}
\label{main_loss}
    \mathcal{L}^{full} = \min_{\theta^{gen}}\max_{\theta^{dom}} \mathcal{L}^{sup}+\lambda_1 \mathcal{L}^{sim} - \lambda_2 \mathcal{L}^{dom},
\end{equation}
where $\theta^{gen}=\{\theta^{opa}, \theta^{sup}, \theta^{sim}\}$ and $\lambda_1, \lambda_2$ are hyper-parameters. 

\subsection{Label Correction}
\label{sec:label_correction}
After training the object placement model, we can update the pseudo rationality labels $\hat{y}^t_{n,k}$ for unlabeled dataset based on the prediction from the latest model. To mitigate the noise issue of pseudo rationality labels, inspired by \cite{Yu_Lin_2023}, we combine the new pseudo rationality labels $\hat{y}^{t'}_{n,k}$ 
with the old pseudo rationality labels  $\hat{y}^{t}_{n,k}$   using a ratio \(\alpha\) as follows,
\begin{equation}
\label{label_correction}
    \hat{y}^{t''}_{n,k} = (1 - \alpha) \cdot \hat{y}^{t'}_{n,k} + \alpha \cdot \hat{y}^{t}_{n,k},
\end{equation}
in which $\hat{y}^{t''}_{n,k}$ is the pseudo label used for the next iteration. 

\section{Unlabeled Dataset Construction}
We construct the unlabeled dataset Open Object Placement Assessment (OOPA) dataset based on Open Images dataset \cite{Kuznetsova_Rom_Alldrin_Uijlings_Krasin_Pont_Tuset_Kamali_Popov_Malloci_Kolesnikov}.

\subsection{Foreground Object Selection}
Open Images dataset \cite{Kuznetsova_Rom_Alldrin_Uijlings_Krasin_Pont_Tuset_Kamali_Popov_Malloci_Kolesnikov} provides instance segmentation masks for 2.8 million object instances in 350 categories. To accommodate our task, we selected 124 object categories from 350 categories, prioritizing those that appear more frequently and independently. Among these selected categories, 39 are overlapped with the object categories in OPA dataset, while the remaining 85 categories are new additions. 

\subsection{Background and Foreground Generation}
By using the segmentation masks, we create pairs of foreground and background in a similar way to prior studies \cite{Zhu_Lin_Cohen_Kuen_Zhang_Chen_2023}. After removing the foreground objects, we adopt the off-the-shelf model of Stable Diffusion V2 \cite{Rombach_2022_CVPR} for inpainting to complete the background images. In total, we obtain over 280,000 pairs of background images and foreground objects covering 124 categories. Nonetheless, a considerable number of foregrounds are either incomplete or occluded by other background objects, which are not suitable for object placement task. To address this issue, we  annotate the label of completeness of $2000$ foreground objects, based on which
a ResNet-50 \cite{He_Zhang_Ren_Sun_2016} classifier is trained to select complete foreground objects.  Finally, we have 146,553 pairs of background images and foreground objects.
Given each foreground-background pair, we randomly sample 30 placements for the foreground object and ensure that the foreground object appears completely in the background. 


\begin{table}[t]
\caption{Comparison of different methods using different training sets on OOPA-s and OOPA-e test sets.}
  \centering
  \resizebox{\columnwidth}{!}{ 
  \begin{tabular}{lcccccc}
    \toprule
     Method & Train Set & F1-s$\uparrow$ & Bal-s$\uparrow$ & F1-e$\uparrow$ & Bal-e$\uparrow$  \\
    \midrule
    SOPA        & OPA          & 0.801 & 0.729 & 0.791 & 0.721  \\
    SOPA        & OPA+OOPA    & 0.846 & 0.806 & 0.808 & 0.740  \\
    Ours(SOPA)    & OPA+OOPA    & \textbf{0.889} & \textbf{0.835} & \textbf{0.840} & \textbf{0.776} \\
    \hline
    FOPA        & OPA          & 0.729 & 0.687 & 0.660 & 0.607 \\
    FOPA        & OPA+OOPA    & 0.793 & 0.765 & 0.778 & 0.682  \\
    Ours(FOPA)    & OPA+OOPA    & \textbf{0.865} & \textbf{0.807} & \textbf{0.852} & \textbf{0.746}  \\
  \bottomrule
  \end{tabular}
  }
\label{tab:OOPA}
\end{table}

\section{Experiment}
\subsection{Experimental Setting}
\subsubsection{Dataset} 
We conduct experiments on OPA dataset \cite{Liu_Zhang_Li_Niu_Liu_Zhang_2021} and our constructed OOPA dataset. OPA dataset includes 62,074 placements for training and 11,396 placements for testing. OPA dataset contains 1,389 different backgrounds scenes and 4,137 different foreground objects from 47 different categories. We split our OOPA dataset to 2,797,081 placements for training and 1,632 placements for testing. For the test set, we manually annotate the rationality labels (1,247 positive and 385 negative).

\subsubsection{Evaluation Metrics} 
The evaluation metrics for object placement task can be divided into discriminative and generative metrics. The discriminative metrics evaluate the performance of predicted rationality labels. Following \cite{Niu_Liu_Liu_Li_2022}, we calculate balanced accuracy and F1-score based on the predicted rationality labels and the ground-truth ones. 

The generative metrics evaluate the quality of composite images obtained by using the predicted rational placements. 
For generative object placement baselines, we use their predicted rational placements to obtain composite images. For discriminative object placement baselines, based on their predicted rationality score maps, we extract top 50 placements with the largest rationality scores, and randomly sample 5 placements from them. Based on the obtained composite images, following \cite{Zhou_Liu_Niu_Zhang_2022}, we adopt accuracy, FID  \cite{Heusel_Ramsauer_Unterthiner_Nessler_Hochreiter_2017}, and LPIPS \cite{Zhang_Isola_Efros_Shechtman_Wang_2018}. Accuracy measures the percentage of generated composite images classified as positive by a binary classifier SimOPA \cite{Liu_Zhang_Li_Niu_Liu_Zhang_2021}.  FID calculates the similarity between generated composite images and positive composite images. 
LPIPS assesses generation diversity by quantifying perceptual differences between pairs of composite images. 
\begin{table}[t]
    \caption{Ablation studies of our framework.}
  \centering
  \resizebox{\columnwidth}{!}{ 
  \begin{tabular}{lcccccccc}
    \toprule
     &\(\mathcal{L}^{sim}\)&\(\mathcal{L}^{dom}\)&label correction&  F1-s$\uparrow$ & Bal-s$\uparrow$& F1-e$\uparrow$ & Bal-e$\uparrow$ \\
    \midrule
    (a)&\(\checkmark\) & &  &0.854 & 0.812 &0.823 & 0.733 \\
    (b)&  &\(\checkmark\) &       &0.848 & 0.810&0.825 & 0.735 \\
    (c)&  & &\(\checkmark\)       &0.873 & 0.825&0.826 & 0.741 \\
    (d)&\(\checkmark\) &\(\checkmark\) &      &0.860 & 0.813 &0.837 & 0.736 \\
    (e)& &\(\checkmark\) &\(\checkmark\)   & 0.875 & 0.830  & 0.827 & 0.748  \\
    (f)&\(\checkmark\) & &\(\checkmark\)     &0.876 & 0.827&0.829 & 0.744\\
    (g)&\(\checkmark\) &\(\checkmark\) &\(\checkmark\)      &\textbf{0.889} & \textbf{0.835} &\textbf{0.840} & \textbf{0.776}\\
  \bottomrule
  \end{tabular}
  }

    \label{tab:ab1}
\end{table}
\begin{table}[t]
\caption{Comparison with the state-of-the-art results on  OPA test set.}
\centering
\resizebox{0.75\columnwidth}{!}{
\begin{tabular}{lccc}
\toprule
Method & \multicolumn{2}{c}{Plausibility} & \multicolumn{1}{c}{Diversity} \\ 
\cmidrule(r){2-3} \cmidrule(l){4-4}
  & Acc.$\uparrow$ & FID$\downarrow$ & LPIPS$\uparrow$ \\ 
\midrule
TERSE \cite{Tripathi_Chandra_Agrawal_Tyagi_Rehg_Chari_2019} & 0.679 & 46.94 & 0 \\
PlaceNet \cite{Zhang_Wen_Min_Wang_Han_Shi_2020} & 0.683 & 36.69 & 0.160 \\
GracoNet \cite{Zhou_Liu_Niu_Zhang_2022} & 0.847 & 27.75 & 0.206 \\
TopNet \cite{Zhu_Lin_Cohen_Kuen_Zhang_Chen_2023} & 0.904 & 25.39 & 0.202 \\
IOPRE \cite{Zhang_Meng_Liu_Nie_Zhong_Fan_Ji} & 0.895 & 21.59 & 0.214 \\
\textbf{Ours} & \textbf{0.957} & \textbf{18.85} & \textbf{0.216} \\
\bottomrule
\end{tabular}
}
\label{tab:FID}
\end{table}
\subsubsection{Implementation Details}
Our method is implemented using PyTorch and distributed on NVIDIA RTX 3090 GPU. We adopt pretrained SOPA \cite{Liu_Zhang_Li_Niu_Liu_Zhang_2021} and FOPA \cite{Niu_Liu_Liu_Li_2022}  as the base models to produce rationality features.  We train our model using Adam optimizer with a learning rate of $5\times10^{-4}$, and reduce the learning rate by a factor of 2 every 2 epochs. We set the ratio \(\alpha\) in Eq. \ref{label_correction} to 0.4, $\lambda_1$ and  $\lambda_2$ in Eq. \ref{main_loss} to 0.5 and 0.1, respectively. Every 25 epochs, we perform label prediction and correction for the unlabeled dataset.
\begin{figure*}[tb]
  \centering
\includegraphics[width=0.91\textwidth]{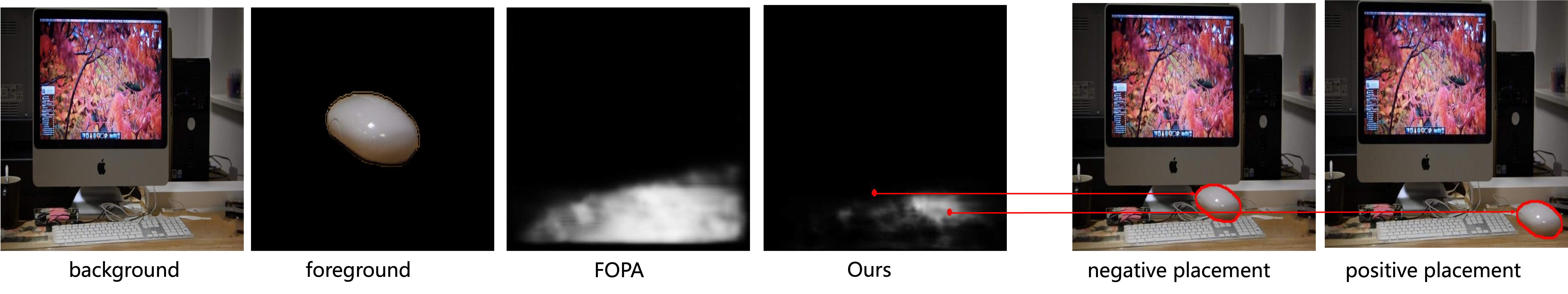}
  \caption{The comparison of the rationality score maps predicted by FOPA model and our model. The foregrounds are highlighted with red outlines.}
  \label{fig:example_heatmap}
\end{figure*}
\subsection{The Effectiveness of Framework}

To demonstrate that our framework can accommodate arbitrary discriminative object placement models, we adopt SOPA \cite{Liu_Zhang_Li_Niu_Liu_Zhang_2021} and FOPA \cite{Niu_Liu_Liu_Li_2022} respectively as the base model to produce rationality features. SOPA takes in a composite image and produce the rationality feature for this placement. FOPA uses background UNet and foreground encoder to produce the rationality features for all candidate placements. 

We train different methods on OPA or OPA+OOPA, and evaluate them on OOPA test set. For baselines trained on OPA+OOPA, we employ label correction in Section~\ref{sec:label_correction}  to leverage unlabeled OOPA. We adopt the discriminative evaluation metrics: F1-score and balanced accuracy. To investigate the benefit of unlabeled dataset and our framework for novel categories, we split OOPA test set into two parts according to the foreground categories. We refer to the part with the foreground categories within (\emph{resp.}, beyond) OPA dataset as OOPA-s (\emph{resp.}, OOPA-e). We also use `-s' and `-e' to distinguish the evaluation results on two parts. 

The experimental results are summarized in Table \ref{tab:OOPA}. It can be seen that on both OOPA-s and OOPA-e test sets, for both object placement models, using unlabeled data can boost the performance and our framework can further significantly improve the results. By comparing the results of SOPA and FOPA on OOPA-e test set using OPA training set, FOPA shows weaker cross-category generalization ability. However, after using unlabeled data and our framework, FOPA can achieve comparable results with SOPA. 

We also present the comparison results of the rationality score maps predicted by FOPA trained on OPA+OOPA and ours(FOPA) on OOPA test set. As shown in Figure \ref{fig:example_heatmap}, our method can provide more reasonable and precise placements, which verifies the generalization ability of our methods to open-set object placements. 

\subsection{Ablation Studies}
\label{Ablation_Studies}
To evaluate the impact of each component in our framework, including \(\mathcal{L}_{sim}\), \(\mathcal{L}_{dom}\), and label correction, we take SOPA as base model and perform ablation studies by individually removing one or two components. Based on Table \ref{tab:ab1}, it is evident that the removal of  \(\mathcal{L}_{sim}\), \(\mathcal{L}_{dom}\), or label correction leads to performance drop on both F1-score and balanced accuracy on both test sets. The optimal result is achieved by the full-fledged framework as shown in row (g).

\begin{figure*}[tb]
  \centering
\includegraphics[width=0.91\textwidth]{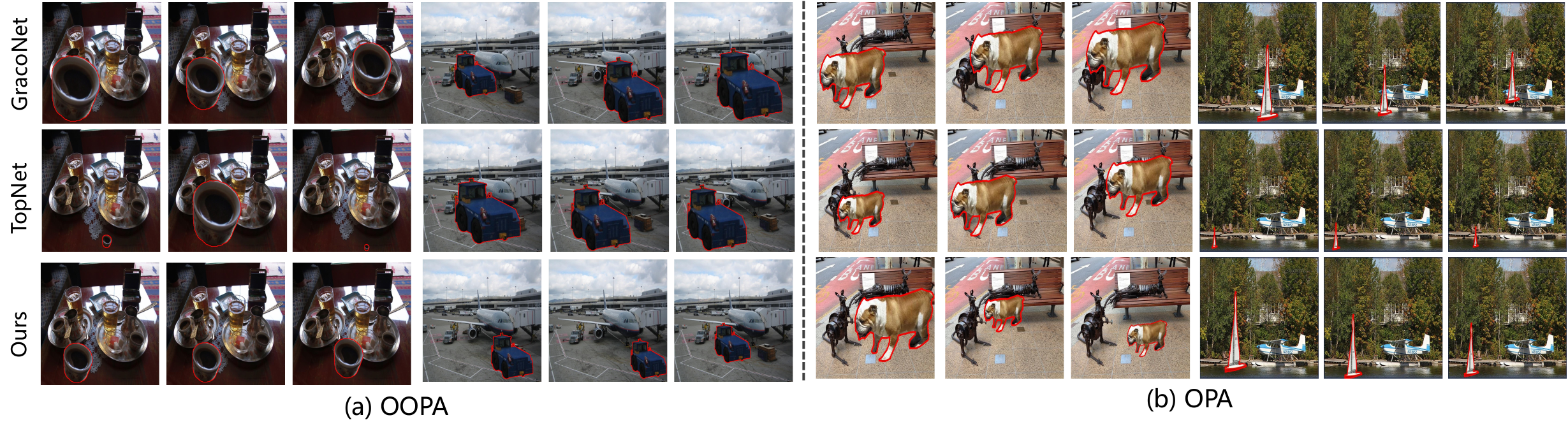}
  \caption{Qualitative comparison with object placement baselines on OPA test set and OOPA test set. The foregrounds are highlighted with red outlines.}
  \label{fig:example_Qual}
\end{figure*}

\subsection{Comparison with the SOTA Results}
On OPA test set, we compare the results obtained by our framework with the results of previous works: TERSE \cite{Tripathi_Chandra_Agrawal_Tyagi_Rehg_Chari_2019}, PlaceNet \cite{Zhang_Wen_Min_Wang_Han_Shi_2020}, GracoNet \cite{Zhou_Liu_Niu_Zhang_2022}, IOPRE \cite{Zhang_Meng_Liu_Nie_Zhong_Fan_Ji}, TopNet \cite{Zhu_Lin_Cohen_Kuen_Zhang_Chen_2023}. 
Following most of previous works \cite{Zhou_Liu_Niu_Zhang_2022,Zhang_Meng_Liu_Nie_Zhong_Fan_Ji}, we report the generative evaluation metrics. The results of TERSE, PlaceNet, GracoNet, IOPRE are directly copied from \cite{Zhou_Liu_Niu_Zhang_2022, Zhang_Meng_Liu_Nie_Zhong_Fan_Ji}. The results of TopNet are reproduced based on the released model. These methods are trained on OPA training set. TERSE \cite{Tripathi_Chandra_Agrawal_Tyagi_Rehg_Chari_2019} can only produce one placement per foreground-background pair, so its diversity score is zero. Our results are obtained using FOPA as base model and OPA+OOPA as training set. As shown in Table \ref{tab:FID}, our results exhibit much better plausibility and reasonable diversity, compared with previous results. The performance gain is attributed to our constructed OOPA dataset and our designed semi-supervised framework. 

\subsection{Qualitative Results and Visualization}
As shown in Figure \ref{fig:example_Qual}, using FOPA as the base model with OPA+OOPA as the training set, we visualize object placement for sample foregrounds in various background scenes on the OPA and OOPA test sets. Compared with baselines, our method avoids unreasonable occlusions and produces more proportionate foreground sizes. 
\section{Conclusion}
In this paper, we have proposed a semi-supervised object placement learning framework which can jointly use small-scale labeled dataset and large-scale unlabeled dataset. Extensive experiments have corroborated that our framework can improve the generalization ability of object placement models, advancing towards open-set object placement.

\section*{Acknowledgment} The work was supported by the National Natural Science Foundation of China (Grant No. 62471287).


\bibliographystyle{IEEEbib}
\bibliography{icme2025references}

\end{document}